\documentclass[final]{cvpr}

\usepackage{times}
\usepackage{epsfig}
\usepackage{graphicx}
\usepackage{amsmath}
\usepackage{amssymb}
\usepackage[square,sort,comma,numbers]{natbib}
\usepackage[T1]{fontenc}

\usepackage[pagebackref=true,breaklinks=true,colorlinks,bookmarks=false]{hyperref}

\newif\ifcomments

\commentsfalse

\ifcomments
\newcommand{\comments}[1]{#1}
\else
\newcommand{\comments}[1]{}
\fi

\newcommand{\rosanne}[1]{\comments{\textcolor{magenta}{[rosanne: #1]}}}

\def\cvprPaperID{10119} 
\def\confYear{CVPR 2021}

\begin{document}

\title{Natural Adversarial Objects}

\author{Felix Lau\\
Scale AI\\
{\tt\small felixlaumon@gmail.com}
\and
Nishant Subramani\\
Allen Institute for AI, Masakhane\\
{\tt\small nishants@allenai.org}
\and
Sasha Harrison\\
Scale AI\\
{\tt\small sasha.harrison@scale.com}
\and
Aerin Kim\\
Scale AI\\
{\tt\small aerin.kim@scale.com}
\and
Elliot Branson\\
Scale AI\\
{\tt\small elliot.branson@scale.com}
\and
Rosanne Liu\\
ML Collective\\
{\tt\small rosanne@mlcollective.org}
}
\maketitle

\begin{abstract}
Although state-of-the-art object detection methods have shown compelling performance, models often are not robust to adversarial attacks and out-of-distribution data.
We introduce a new dataset, Natural Adversarial Objects (NAO), to evaluate the robustness of object detection models. NAO contains 7,934 images and 9,943 objects that are unmodified and representative of real-world scenarios, but cause state-of-the-art detection models to misclassify with high confidence. 
The mean average precision (mAP) of EfficientDet-D7 drops 74.5\% when evaluated on NAO compared to the standard MSCOCO validation set.

Moreover, by comparing a variety of object detection architectures, we find that better performance on MSCOCO validation set does not necessarily translate to better performance on NAO, suggesting that robustness cannot be simply achieved by training a more accurate model.

We further investigate why examples in NAO are difficult to detect and classify. Experiments of shuffling image patches reveal that models are overly sensitive to local texture. Additionally, using integrated gradients and background replacement, we find that the detection model is reliant on pixel information within the bounding box, and insensitive to the background context when predicting class labels. NAO can be downloaded \href{https://drive.google.com/drive/folders/15P8sOWoJku6SSEiHLEts86ORfytGezi8?usp=sharing}{here}.
\end{abstract}

\section{Introduction}

\begin{figure}[t]
\begin{center}
   \includegraphics[width=1.0\linewidth]{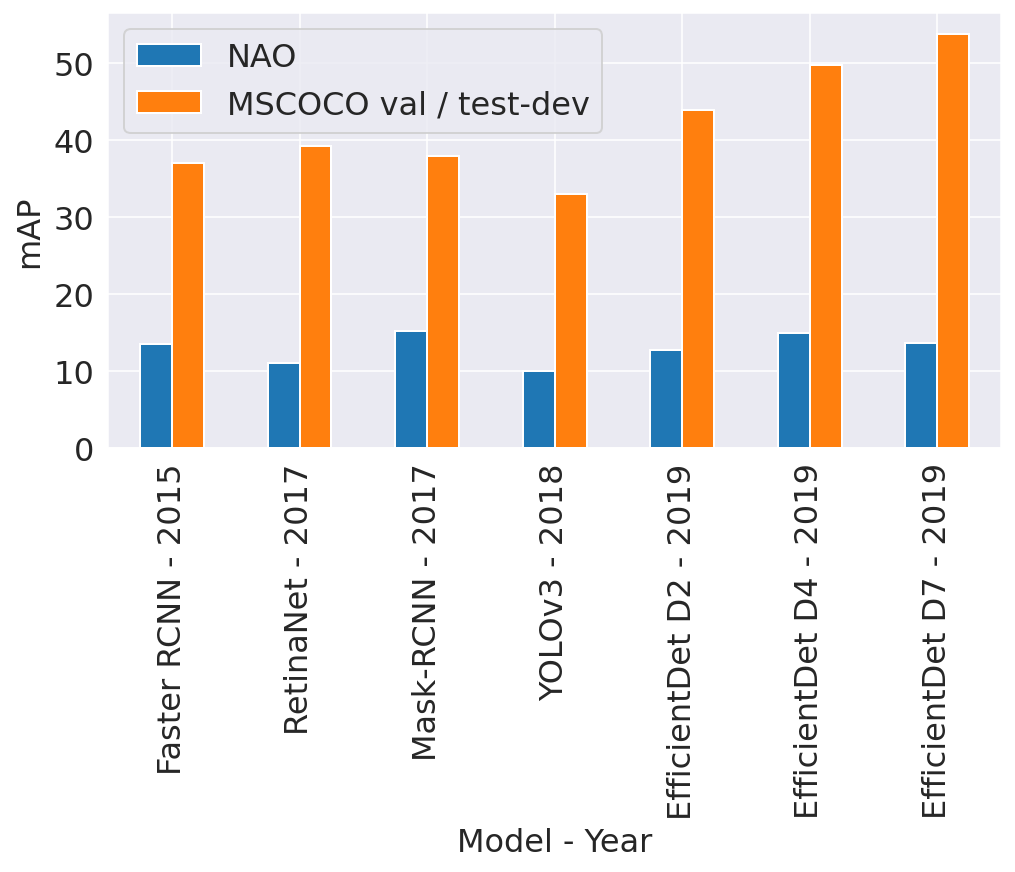}
\end{center}
   \caption{Mean average precision (mAP) of various detection models evaluated on NAO and MSCOCO \textit{val} or \textit{test-dev} set. All models show significant reduction in performance on NAO despite their accuracy improvement in MSCOCO in recent years. NAO is a challenging test set for detection models trained on MSCOCO and future work is required to close the performance gap.}
\label{fig:long}
\label{fig:onecol}
\label{fig:evaluation_results}
\end{figure}

It is no longer surprising to have machine learning vision models perform well on large scale training sets and also generalize on canonical test sets coming from the same distribution. However, generalization towards difficult, out-of-distribution samples still poses difficulty. \rosanne{consider remove: In real world applications, both the frequency and severity of misclassifications contribute to the overall quality of a machine learning system.} Recht \etal~\cite{Recht2019-pm} showed that model performance on canonical test sets is an overestimate of how they will perform on new data. Moreover, recent research on adversarial attacks has shown that deep neural networks are surprisingly vulnerable to artificially manipulated images, casting new doubt on the efficacy and security of such models.
 
 The vulnerability of neural networks to adversarial attacks that are deliberately generated to fool the system is unsurprising, and well studied. However,
 \rosanne{Not really relevant to intro L_p attacks here. Consider removing: 
 The type of attack that leverages artificial images generated to fool a specific classifier is commonly referred to as an $L_p$ adversarial attack.
Formally, this type of attack creates a synthetic image $x_{adv}$ with the SGD objective of maximizing classification loss, subject to the constraint $||x_{adv} - x ||_p < \epsilon $, where $ || \cdot ||_p$ denotes the $L_p$ norm. This $L_p$ norm constraint enforces that the synthesized image is "close" to the original, such that the difference is imperceptible to the human eye, but results in completely different predictions for a deep learning model~\cite{gilmer2018motivating}.
The vulnerability of neural networks to $L_p$ attacks is well documented, yet}
this type of attack represents a narrow threat model because it necessitates that the adversary has control over the raw input, or has access to the model weights. It is often overlooked that real-world, unmodified images can also be used adversarially to cause models to fail. These ``natural" adversarial attacks represent a less restricted threat model, where an attacker can easily create black-box attacks without carefully constructing input perturbations~\cite{gilmer2018motivating}, but only by using  naturally occurring images that are easily obtainable. Such images are called natural adversarial examples~\cite{Hendrycks2019-cn}: unmodified, real-world images that cause modern image classification models to make egregious, high-confidence errors.

In~\cite{Hendrycks2019-cn} natural adversarial examples are only constructed for image classification models. In this work, we seek to create an evaluation set analogous to~\cite{Hendrycks2019-cn}, but instead targeted at object detection tasks. We name such a dataset Natural Adversarial Objects (NAO). The goal of NAO is to benchmark the worst case performance of state-of-the-art object detection models, while requiring that examples included in the benchmark are unmodified and naturally occurring in the real world. 

We present a method to identify natural adversarial objects using a combination of existing object detection models and human annotators.
First, we compare the predictions from various off-the-shelf detection models against a dataset already annotated with ground truth bounding boxes. We consider images containing high confidence false positives and misclassified objects as candidates for NAO.
Then, we use a human annotation pipeline to filter out mislabeled images and non-obvious objects (e.g. occluded or blurry objects). Finally, we re-annotate the images using the object categories of the Microsoft Common Objects in Context (MSCOCO) dataset~\cite{lin2014microsoft}.

We perform extensive analyses to understand why objects in NAO are naturally adversarial. We visualize the embedding space common to MSCOCO, OpenImages, and NAO, and show that NAO images exist in the "blind spots" of the MSCOCO dataset. Next, by comparing integrated gradients~\cite{Sundararajan2017-jl} with predicted bounding boxes and replacing object backgrounds, we show that the detection model seldom makes use of object contexts. Lastly, by shuffling patches within the bounding box, we show that models relies on object subparts and texture to detect and classify the objects.

\begin{figure*}[t]
\begin{center}
   \includegraphics[width=1.0\linewidth]{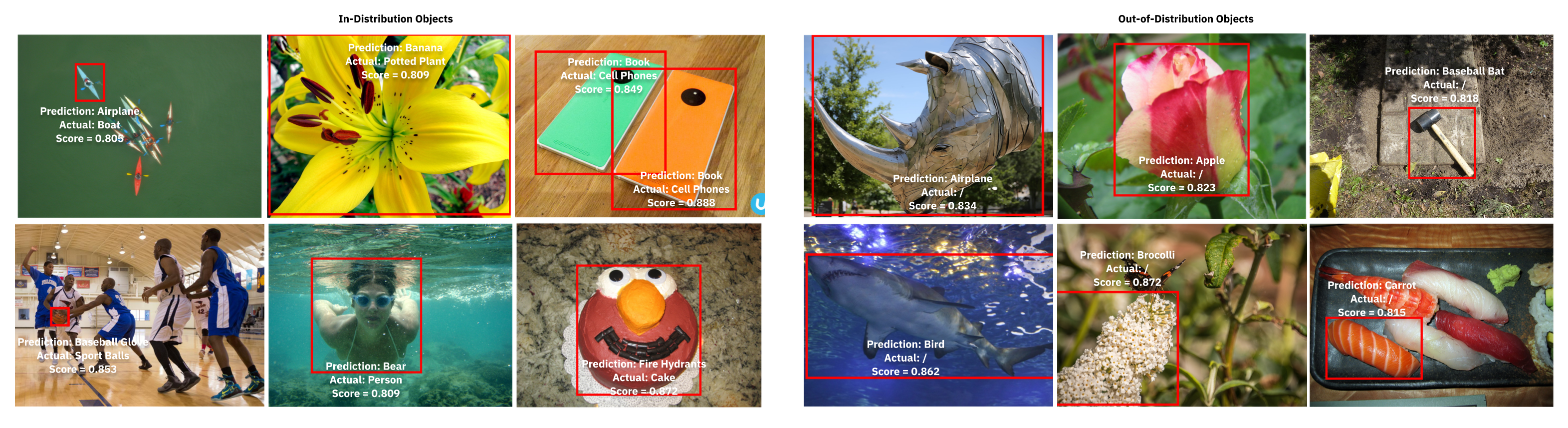}
\end{center}
   \caption{Sample images from NAO where EfficientDet-D7 produces high confidence false positives and egregious classification. \textbf{Left:} High confidence misclassified objects where the ground truth label is in-distribution and among the MSCOCO object categories. \textbf{Right:} High confidence false positives where the ground truth object is out-of-distribution (i.e. not part of MSCOCO object categories). The misclassified objects and false positives are superficially similar to the predicted classes -- for example, the fin of the shark is visually similar to the airplane tail and the yellow petals of the flower is similar to a bunch of bananas.}
\label{fig:long}
\label{fig:onecol}
\label{fig:samples}
\end{figure*}

\section{Related Work}

\paragraph{Natural Adversarial Examples} 

Hendrycks \etal~\cite{Hendrycks2019-cn} construct two datasets, namely \textit{ImageNet-A} and \textit{ImageNet-O}, to measure the robustness of image classifiers against out of distribution examples.  To construct these two datasets, they choose images on which a pretrained ResNet model failed to make a correct prediction. We adopt a similar approach for selecting adversarial examples but use an object detection model and take extra steps to ensure high quality annotations by using human annotators.

\paragraph{Adversarial Examples}

Adversarial examples are inputs that are specifically designed to cause the target model to produce erroneous outputs. 
Since the findings of Szegedy \etal~\cite{Szegedy2013-jl}, there has been a flurry of research addressing adversarial attacks. Defensive distillation ~\cite{DBLP:journals/corr/PapernotMWJS15}, one of the promising defense mechanisms against adversarial examples, was invented but then defeated within a year by a novel attack method proposed by Carlini \etal~\cite{7958570}. Similarly, defenses using adversarial training~\cite{46638}, once considered to be robust to whitebox attacks, along with other methods that rely on gradient masking, such as Defense GAN~\cite{samangouei2018defensegan}, stochastic activation pruning~\cite{s.2018stochastic}, Pixeldefend~\cite{song2018pixeldefend}, etc., are now each proven to be vulnerable to other types of attacks. These developments demonstrate how difficult it is to successfully defend against adversarial examples.

There are many hypotheses as to why adversarial examples pose such a challenge for classifiers. Papernot \etal~\cite{DBLP:journals/corr/PapernotMG16} suggested that adversarial examples cause models to fail because they lie in the low probability region of the data manifold. Goodfellow \etal~\cite{Goodfellow2014-rh} showed that deep neural networks are especially vulnerable to adversarial examples due to the local linearity property, and proposed FGSM to fool the deep neural networks. Although deep neural networks use non-linear activation functions, we often observe that models operate in the linear regions of the activation functions to avoid the vanishing gradient problem~\cite{hochreiter98}. 

Arpit \etal~\cite{pmlr-v70-arpit17a} analyzed the capacity of neural networks to memorize training data, and found that models with a high degree of memorization are more vulnerable to adversarial examples. Jo \etal~\cite{DBLP:journals/corr/abs-1711-11561} have shown that convolutional neural networks tend to learn the statistical regularities in the training dataset, rather than the high level abstract concepts. Since adversarial examples are transferable between models that are trained on the same dataset, these different models may have learned the same statistics and therefore are vulnerable to similar adversarial attacks. Brendel \etal~\cite{brendel2018approximating} show that small local image features are sufficient for deep learning model to achieve high accuracy. Geirhos \etal~\cite{geirhos2018imagenettrained} show that ImageNet-trained CNNs are biased toward texture and created \textit{Stylized-ImageNet} to reveal the severity of such bias. Similarly, Ilyas \etal~\cite{ilyas2019adversarial} showed that adversarial examples are a byproduct of exploiting non-robust features that exist in a dataset. Non-robust features are derived from patterns in the data distribution that are highly predictive, yet brittle and incomprehensible to humans. Undoubtedly, the reasons behind the existence and pervasiveness of adversarial examples still remains an open research problem.



\paragraph{Model Interpretability} While the interpretability of deep neural networks remains an open research question, there exist attribution methods that help explain the relationships between the input and output of such models. In simpler terms, they can be used to understand why a model makes mistakes. Sundararajan \etal~\cite{Sundararajan2017-jl} suggests that attribution methods should satisfy two axioms: sensitivity and implementation invariance, and proposes a new method, \textit{Integrated Gradient}, to understand which parts of an image influence the prediction the most.

\paragraph{Object Detection Architectures} Detection models fall into two categories:  one-stage (\cite{tan2020efficientdet},~\cite{Redmon2018-db},~\cite{redmon2017yolo9000}) and two-stage models (\cite{ren2015faster},~\cite{Sermanet2013-ob}), differentiated by whether the model has a region pooling stage. Single-stage model are more computationally efficient, but usually less accurate than the 2-stage models. In this paper, we evaluate both single and two-stage models using the NAO dataset. Tan \etal~\cite{tan2020efficientdet} introduced EfficientDet, which uses EfficientNet~\cite{Tan2019-sq} as backbone and uses BiFPN such that the model is more efficient while more accurate, achieving state-of-the-art results in MSCOCO at 54.4 on the \textit{val} set.



\section{Creating Natural Adversarial Objects (NAO) Dataset}

\subsection{Limitations of MSCOCO} \label{subsection:limitation_of_mscoco}
MSCOCO~\cite{lin2014microsoft} is a common benchmark dataset for object detection models.
It contains 118,287 images in the training set, 5,000 images in the \textit{val} set and 20,288 images in the \textit{test-dev} set. MSCOCO contains 80 object categories consisting of common objects such as \textit{horse}, \textit{clock}, and \textit{car}.
The goal of MSCOCO is to introduce a large-scale dataset that contains objects in non-iconic or non-canonical views.
The images in MSCOCO were originally sourced from Flickr, then filtered down in order to limit the scope of the benchmark to a set of 80 categories.
These 80 categories were chosen from a list of the most commonplace visually identifiable objects.  Still, this category list represents only a small subset of object categories in real life.
For example, 'fish' is not among the 80 categories, and as a result there are only a few photos taken underwater.
This leads to a biased benchmark with limitations for generalizability and robustness.
As a result, in this work, we ensure more diverse sourcing --- choosing images from OpenImages v6~\cite{Kuznetsova2018-bk}, a dataset with 600 object categories, in order to create a more representative dataset.

\subsection{Sourcing Images for NAO from OpenImages}
To create NAO, we first sourced images from the training set of OpenImages~\cite{Kuznetsova2018-bk}, a large, annotated image dataset containing approximately 1.9 million images and 15.8 million bounding boxes across 600 object classes.

One challenge of using OpenImages is that the bounding boxes are not exhaustively annotated. Each image is first annotated with positive and negative labels which indicate the presence or absence of an object in the image.
Only objects belonging to the positive label categories are annotated with bounding boxes.
As a result, some objects that belong to the OpenImages object categories are not labeled with a bounding box. 
For example, imagine both horse and pig are represented in the 600 object classes. If an image contains a horse and a pig, and only the category of horse is included in preliminary round of positive labels, then the image would be labeled with a bounding box for the horse but not the pig.
This non-exhaustive annotation approach makes it difficult to produce and compare precision and recall to other exhaustively annotated dataset such as MSCOCO. This is because false positives and false negatives can only be evaluated accurately if the ground truth bounding boxes are exhaustive. 

One other challenge that arises when sourcing images from OpenImages is that the object categories of OpenImages and MSCOCO are not the same. Therefore, after obtaining a set of natural adversarial images, we exhaustively annotate the images with all 80 MSCOCO object classes to facilitate straightforward comparison between NAO and the MSCOCO \textit{val} and \textit{test} sets.

\subsection{Candidate Generation}
To generate object candidates, we perform inference on OpenImages using an EfficientDet-D7 model ~\cite{tan2020efficientdet} pretrained on MSCOCO, which yields predicted object bounding boxes for each candidate image.
Our goal is to find two types of errors: (i) \textbf{hard false positives} (i.e. false positives with high confidence) and (ii) \textbf{egregiously misclassified} objects.  
For a detection to be a hard false positive, we require the prediction to have no matching ground truth box with intersection over union (IoU) greater than 0.5, but to have a class confidence greater than 0.8.
We define egregiously misclassified objects as predictions that have a matching ground truth bounding box with an IoU greater than 0.5, but have an incorrect classification with a confidence greater than 0.8. We do not consider false negatives with high confidence because we observe that these are commonplace especially in crowded scenes.
There are 43,860 images containing at least one hard false positive or egregiously misclassified object.


\begin{table*}
\begin{center}
\begin{tabular}{l|l|l|l|l|l|}
\cline{2-6}
                                      & \multicolumn{2}{c|}{Statistics}      & \multicolumn{3}{c|}{Top 3 Objects (Count)}    \\ \cline{2-6} 
                                      & Number of Images & Number of Objects & 1st             & 2nd         & 3rd           \\ \hline
\multicolumn{1}{|l|}{MSCOCO \textit{val}}      & 5,000            & 36,781            & Person (11,004) & Car (1,932) & Chair (1,791) \\ \cline{1-1}
\multicolumn{1}{|l|}{MSCOCO \textit{test-dev}} & 20,288           & -                 & -               & -           & -             \\ \cline{1-1}
\multicolumn{1}{|l|}{NAO}             & 7,934            & 9,943             & Person (3,551)  & Cup (1,366) & Car (707)     \\ \cline{1-1}
\hline
\end{tabular}
\end{center}
\caption{Dataset statistics of MSCOCO \textit{val}, \textit{test-dev} and NAO.}
\end{table*}

\begin{figure}[t]
\begin{center}
    \includegraphics[width=1.0\linewidth]{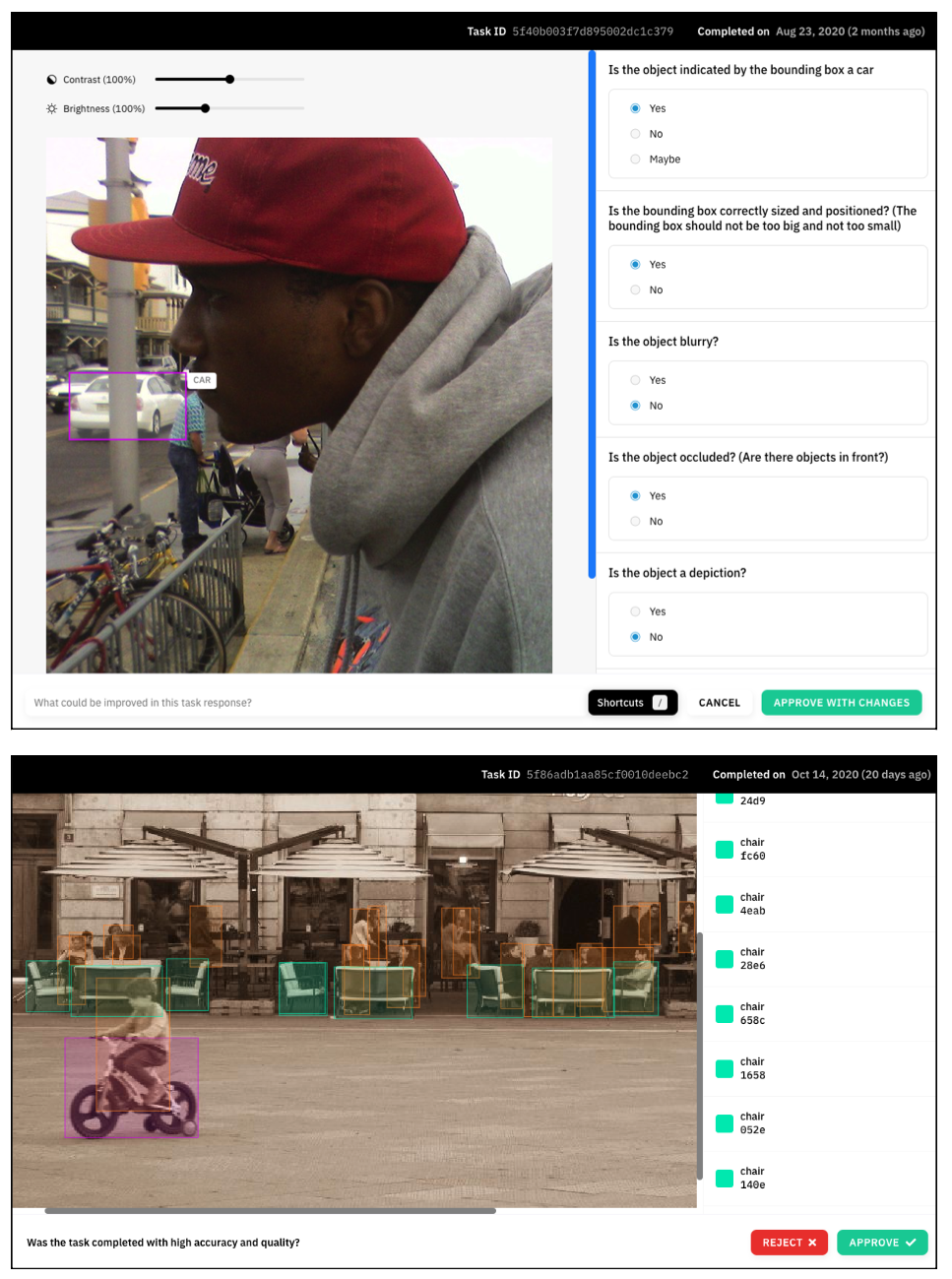}
\end{center}
    \caption{\textbf{Top}: Annotation interface for the first annotation stage (classification) where the annotator confirms that the object belongs to the correct category, not occluded, not blurry and not a depiction.
    \textbf{Bottom}: Annotation interface for second annotation stage (bounding box) where the annotators locate and classify all objects in the images using the MSCOCO object categories.}
\label{fig:long}
\label{fig:onecol}
\label{fig:annotation_interface}
\end{figure}

\subsection{Annotation Process}

Our annotation process has two annotation stages: classification and bounding box annotation.

\paragraph{Classification Stage}
In the classification stage, annotators identify whether the object described by the bounding box shown indeed belongs to the ground truth class as defined by the annotation in OpenImages or as predicted by the EfficientDet-D7.
The purpose of this stage is to remove the possibility that the model prediction is "incorrect" due to the ground truth label being incorrect.

In addition, we ask the annotators to confirm whether the object can be "obviously classified" according to the following criteria: 

\begin{enumerate}
    \item Is the bounding box around the object correctly sized and positioned such that it is not too big or too small? 
    \item Does the object appear blurry? 
    \item Is the object occluded (i.e. are there other objects in front of this one)?
    \item Is the object a depiction of the correct class (such as a drawing or an image on a billboard)?
\end{enumerate}

We ask these additional questions to filter out ambiguous objects, such that a human can easily identify what class an object belongs to.
After this filtering, 18.1\% of the images (7,934) remain; each of the remaining images are confirmed to fulfill the 4 criteria, and represent true misclassifivations by the model.
In this first annotation stage (classification), 5 different annotators are asked to annotate the same image and we use their consensus to produce an aggregated response by majority vote.

\paragraph{Bounding Box Stage}
In the second annotation stage (bounding box), annotators exhaustively identify and put boxes around all objects that belong to the MSCOCO object categories. We are unable to directly use the annotations from OpenImages because there is not a one-to-one mapping between the OpenImages and MSCOCO object categories, and because the bounding box annotations from OpenImages are not exhaustively annotated. However, the bounding box annotations from OpenImages are provided to the annotators as a starting point.

These bounding box annotation tasks are completed by 2 sets of annotators. The first set of annotators complete the bulk of the task by placing bounding boxes around objects that belong to the MSCOCO object categories. The second set of annotators review the work of the first set of annotators, sometimes adding missing bounding boxes or editing the existing ones.

To ensure the quality of the annotation is high, in both of these stages, the annotators have to pass multiple quizzes before they can start working tasks to ensure they understand the instructions well. If the annotator fails to maintain a good score, they are no longer eligible to continue to annotate the images.  This process of vetting annotators is consistent with the methodology used to construct MSCOCO ~\cite{lin2014microsoft}.

When the annotators from the 2 different stages disagree, we tie break by choosing second annotator who is positioned as the reviewer.



\subsection{Evaluation Protocol}

The goal of NAO is to test the robustness of object detection models against edge cases and out-of-distribution images. We propose two main evaluation metrics: \textbf{overall mAP} and \textbf{mAP without out-of-distribution objects}. mAP without out-of-distribution objects evaluates against edge cases of object categories that the detection models are trained on, while the overall mAP evaluates robustness against out-of-distribution objects. For calculating mAP without out-of-distribution objects, any detection matched to an object not belong to the 80 MSCOCO object categories is not considered a false positive.

NAO should be mainly used as a test set to evaluate detection models trained on MSCOCO. However, a split of train, validation and test set is also provided for robustness approaches that require training.


\section{Results}

\subsection{Evaluation of Detection Models}

Figure \ref{fig:evaluation_results} and Table \ref{table:evaluation_results} show the mean average precision (mAP) of several state-of-the-art detection models evaluated on MSCOCO and NAO. Despite the fact that the images in NAO were chosen using an EfficientDet-D7  model, we observe that other object-detection architectures show a similar reduction in mAP when evaluated on NAO. Concretely, when using NAO the mAP of EfficientDet-D7 is reduced by 74.5\%, while Faster RCNN is reduced by 36.3\% when compared to MSCOCO. Even though EfficientDet-D7 was developed more recently than Faster RCNN, the mAP on NAO is similar. This indicates that latest models are not more robust on NAO, despite their superior performance on MSCOCO evaluation sets.  This in turn suggests that modeling improvements from recent years do not address the issue of high confidence misclassification in out-of-distribution samples.

We also calculate the mAP without out-of-distribution objects. That is, if a detection matches a bounding box that does not belong the MSCOCO object categories, the detection is not counted as a false positive. We can see that, this exclusion improves mAPs on NAO, but overall, the results are still considerably worse than those from the MSCOCO \textit{val} and \textit{test-dev} set.


\begin{table*}
\begin{center}
\begin{tabular}{l|l|l|l|l|l|l|}
\cline{2-7}
                                      &        & \multicolumn{1}{c|}{MSCOCO \textit{val}} & MSCOCO \textit{test-dev} & \multicolumn{3}{c|}{NAO}                  \\ \cline{2-7} 
                                      & Params & mAP                             & mAP             & mAP  & mAR  & mAP w/o out-of-distribution \\ \hline
\multicolumn{1}{|l|}{Faster RCNN}     & 42M    & 21.2                            & 21.5            & 13.5 & 41.4 & 22.8                        \\ \cline{1-1}
\multicolumn{1}{|l|}{RetinaNet-R50}   & 34M    & 39.2                            & 39.2            & 11.1 & 37.2 & 19.5                        \\ \cline{1-1}
\multicolumn{1}{|l|}{YOLOv3}          & 62M    & -                               & 33.0            & 10.0 & 28.4 & 17.5                        \\ \cline{1-1}
\multicolumn{1}{|l|}{Mask RCNN R50}   & 44M    & 37.9                            & -               & 15.2 & 43.8 & 24.6                        \\ \cline{1-1}
\multicolumn{1}{|l|}{EfficientDet-D2} & 8.1M   & 43.5                            & 43.9            & 12.8 & 40.2 & 25.4                        \\ \cline{1-1}
\multicolumn{1}{|l|}{EfficientDet-D4} & 21M    & 49.3                            & 49.7            & 15.0 & 42.7 & 29.6                        \\ \cline{1-1}
\multicolumn{1}{|l|}{EfficientDet-D7} & 52M    & 53.4                            & 53.7            & 13.6 & 40.8 & 26.6                        \\ \cline{1-1}
\hline
\end{tabular}
\end{center}
\caption{mAP of various detection models evaluated on MSCOCO \textit{val} and \textit{test-dev} set and NAO. Accuracy of all models were significantly lower on NAO than on MSCOCO. There is a slight increase in mAP when out-of-distribution objects are excluded.}
\label{table:evaluation_results}
\end{table*}

\subsection{Common Failure Modes}

\begin{figure*}[t]
\begin{center}
   \includegraphics[width=1.0\linewidth]{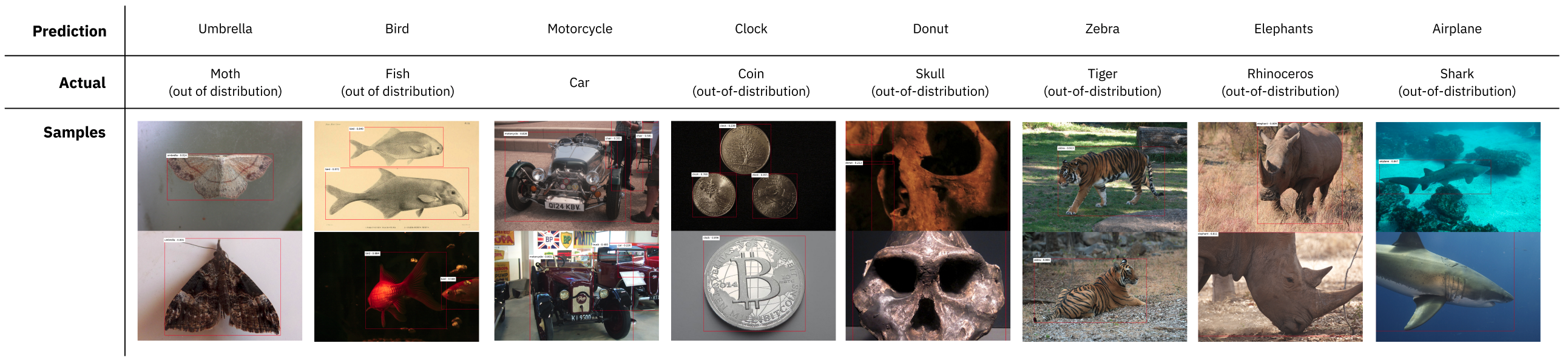}
\end{center}
   \caption{Selected samples to showcase common failure modes.}
\label{fig:long}
\label{fig:onecol}
\label{fig:common_failure_mode}
\end{figure*}

In Figure \ref{fig:common_failure_mode}, we visualize some failure modes of the detection models on NAO. In most of the misclassified objects, the predicted class is superficially similar to the ground truth class, but obviously different in terms of function. For example, clocks and coins are similar in shape (circular), texture (metallic in some cases) and both have characters near the perimeters. However, they are very different in function and in scale, such that any human can easily tell the difference between the two.
Similarly, airplanes and sharks are similar in overall shape, color, and texture, but exist in rather different scenes.

Another common failure mode is differentiating different animal species. For example, elephant and rhinoceros both have somewhat similar skin color and texture but they are very different in size and rhinoceros do not have the distinctive elephant trunk.

\subsection{Dataset Blind Spot}

\begin{figure*}[t]
\begin{center}
   \includegraphics[width=0.95\linewidth]{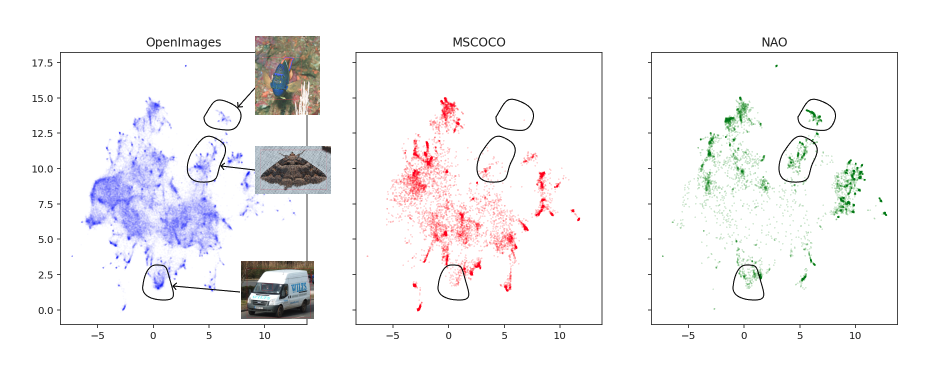}
\end{center}
   \caption{BiT ResNet-50 embeddings projected by UMAP on OpenImages \textit{train}, MSCOCO \textit{train} and NAO. NAO images are under-represented in MSCOCO.}
\label{fig:long}
\label{fig:onecol}
\label{fig:bit_embeddings}
\end{figure*}

As mentioned in Section \ref{subsection:limitation_of_mscoco}, MSCOCO sourced images from Flickr search queries related to the 80 object categories. This process can be seen as a biased sampling process of all captured photos, resulting in "blind spots" in MSCOCO. For example, because there is not any "fish" category, the frequency of photos taken underwater in MSCOCO is much lower than all captured photos. In this section, we investigate this sampling bias by comparing the image embeddings of BiT ResNet-50~\cite{Kolesnikov2019-qu} pretrained on ImageNet-21k~\cite{russakovsky2015imagenet} across the 3 datasets -- OpenImages \textit{train}, MSCOCO \textit{train} and NAO. We consider OpenImages as a proxy for all captured images and with MSCOCO and NAO being a subset of the captured images. The image embedding is the output of the global average pooling layer, resulting in a vector of size 2,048. We then use UMAP~\cite{McInnes2018-ph} to reduce the dimension to 2 for visualization as shown in Figure \ref{fig:bit_embeddings}.

When comparing the embedding space of MSCOCO with OpenImages, we found that there are regions where the density is significantly lower in MSCOCO than in OpenImages. Some of these low-density regions are indicated by the black circles in Figure \ref{fig:bit_embeddings}. When cross-referencing with the embedding space of NAO, we can see that these low-density regions of MSCOCO are in fact high-density in NAO, indicating that the examples in NAO are exploiting the under-represented regions that arise from MSCOCO's biased sampling process. We visualize 3 of such low-density clusters and they each reveal a common failure mode (i.e. fish misclassified as bird, insects misclassified as umbrella and van misclassified as truck.)

\subsection{Background Cues}

Hendrycks \etal\cite{Hendrycks2019-cn} suggest that classification models are vulnerable to natural adversarial examples because classifiers are trained to associate the entire image with an object class, resulting in frequently appearing background elements being associated with a class. Object detection models are different from image classifiers in that they receive additional supervision about the object position and size. We instead argue that the primary cause of detection models being vulnerable to NAO is their tendency to focus too much on the information within the predicted bounding boxes.

In this section, we study the effect of object background on classification probabilities.  Specifically, we quantify the change in probability of the detected object when its background is replaced. We use a MSCOCO-pretrained Mask-RCNN~\cite{he2017mask} with a ResNet 50 backbone to obtain instance segmentation masks on MSCOCO \textit{val} and NAO. Then, we use the instance segmentation masks to retain only the most confident object and replace the rest of the image with a new background. There are 6 new backgrounds -- underwater, beach, forest, road, mountain and sky -- where Mask-RCNN detects no objects of probability higher than 0.1 from the backgrounds themselves. We measure the change of probability by matching the bounding box detected on the original image and the bounding box detected on the new image with the background replaced. We repeat this process for all images in NAO and MSCOCO \textit{val} set and all 6 backgrounds.

\begin{figure}[t]
\begin{center}
   \includegraphics[width=1.0\linewidth]{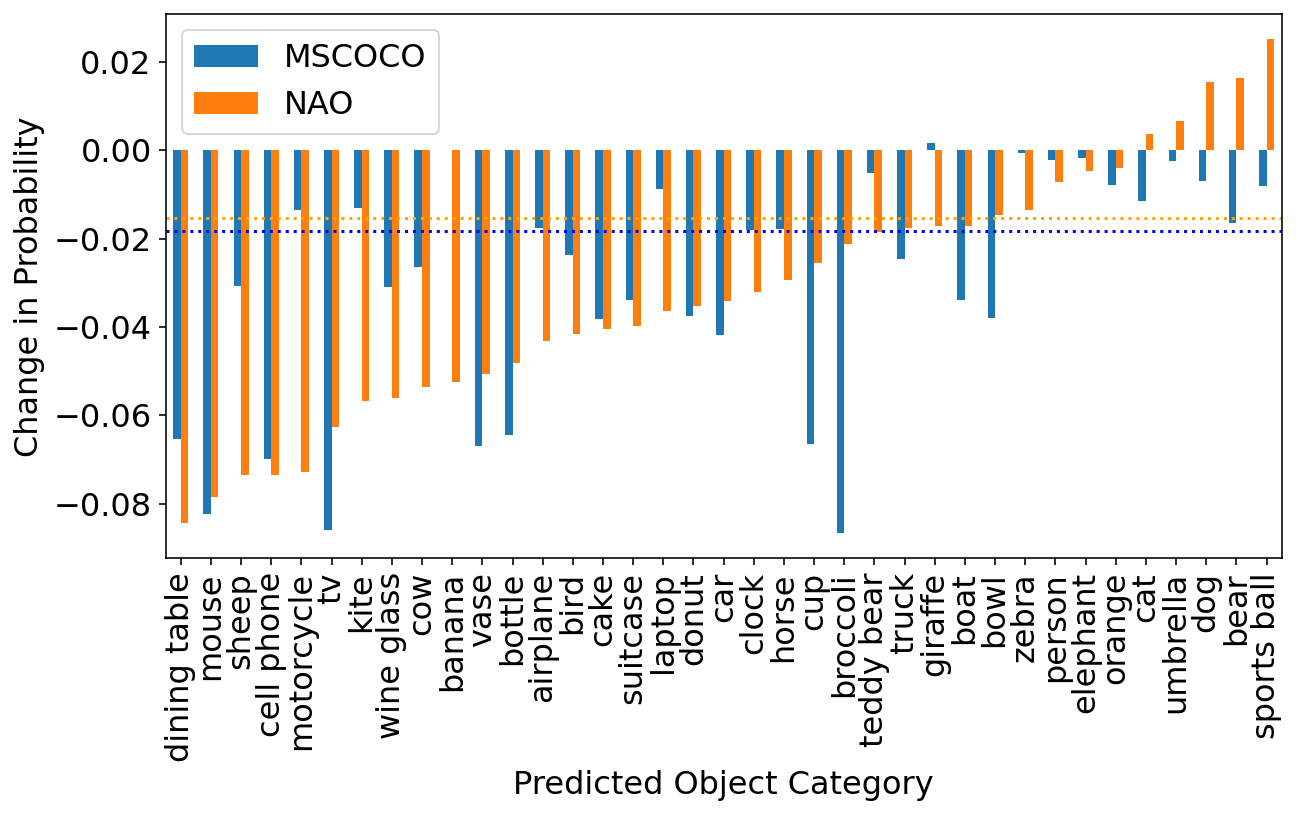}
\end{center}
   \caption{Average change in probability of objects when the backgrounds are replaced. The orange and blue dotted lines indicate average change in probabilities across all classes in MSCOCO and NAO. The small change in probability indicates that the detection model did not make use of background to classify the objects.}
\label{fig:long}
\label{fig:onecol}
\label{fig:background_swap_changes}
\end{figure}

As show in Figure \ref{fig:background_swap_changes}, in both NAO and MSCOCO, the change in probabilities is low, indicating that the model does not make use of the background when detecting the object. While this robustness against background change is favorable in most cases, this also shows that the model also does not account for unlikely combinations of background and foreground objects. For example, when the model misclassifies a shark as an airplane, the network could have noticed that the detected "airplane" is underwater and assigned a lower probability to the class airplane.


\subsection{Integrated Gradients Analysis}

We further try to understand the source of the egregious misclassifications by computing the integrated gradients~\cite{Sundararajan2017-jl} of the network classification head output with respect to the input image. We aim to find the proportion of integrated output within the bounding box to understand if the network makes use the context of the object for detection and classification.




Specifically, we computed the gradients of the classification output of highest-scored bounding box with respect to the input image and measure the proportion of the sum of attribution inside the bounding box with respect to the total attribution. 
When there are multiple same-class objects to detect, we make sure to attribute each object separately. 
For example, when there are 2 people, we calculate the attribution of one person, ensuring the attribution of the other person is not counted towards the background.
We used EfficientDet-D4 and randomly sampled 1000 images for this experiment. We found that for most classes, the majority of the attributions come from inside the bounding box.



\begin{figure}[t]
\begin{center}
   \includegraphics[width=1.0\linewidth]{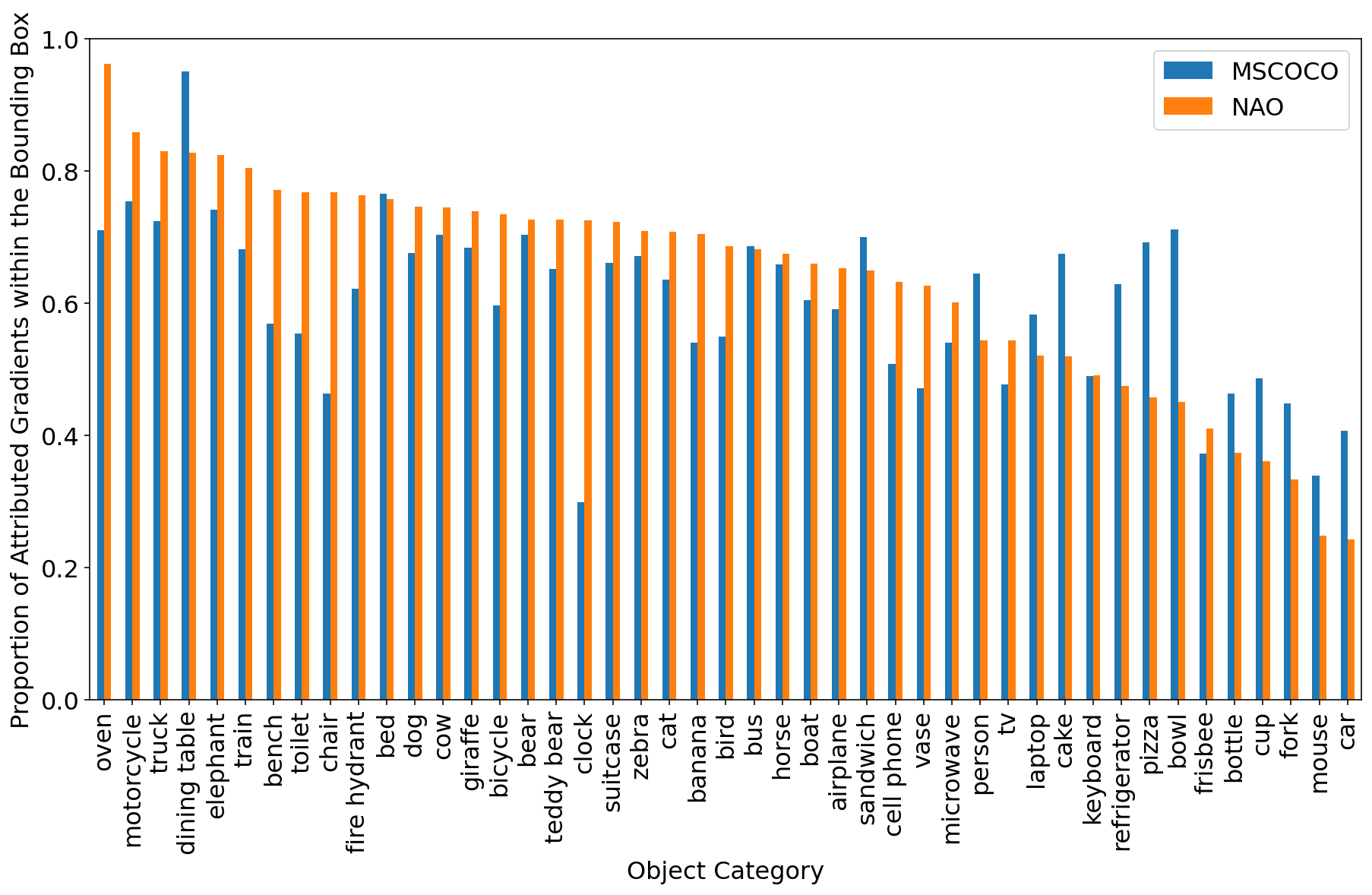}
\end{center}
   \caption{Proportion of attributed gradients within the bounding box by object category. In many classes, the detection model seldom make use of the object surroundings for classification and detection.}
\label{fig:long}
\label{fig:onecol}
\label{fig:integrated_gradients_boxplot}
\end{figure}

Both Figure \ref{fig:background_swap_changes} and Figure \ref{fig:integrated_gradients_boxplot} suggest that the detection model do not make use of background enough and instead mainly focus on the information within the predicted bounding box.


\begin{figure}[t]
\begin{center}
  \includegraphics[width=1.0\linewidth]{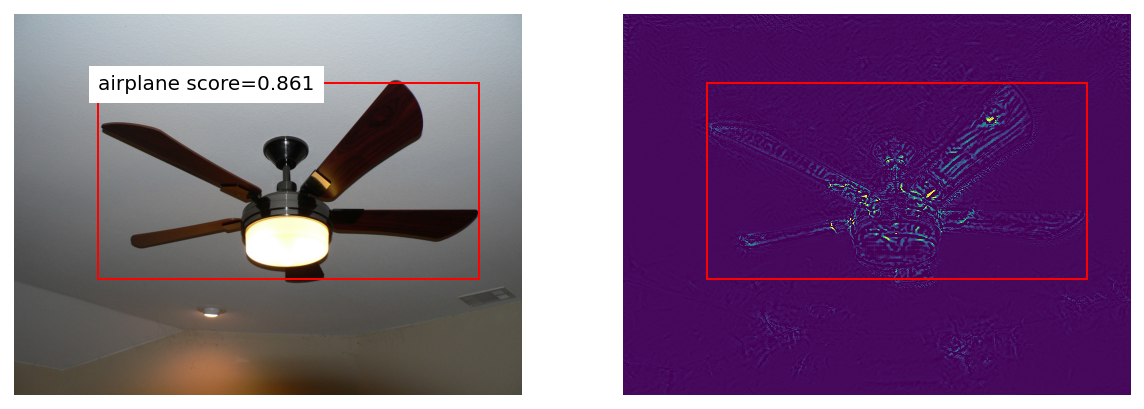}
\end{center}
  \caption{Integrated gradients of a ceiling fan misclassified as an airplane. From the attribution heatmap in the right, we can see that the model focus on the fins of the fan but not the lights in the middle or the fact that the fan is indoors.}
\label{fig:long}
\label{fig:onecol}
\label{fig:integrated_gradients_samples}
\end{figure}


\subsection{Patch Shuffling and Local Texture Bias}

\begin{figure}[t]
\begin{center}
   \includegraphics[width=1.0\linewidth]{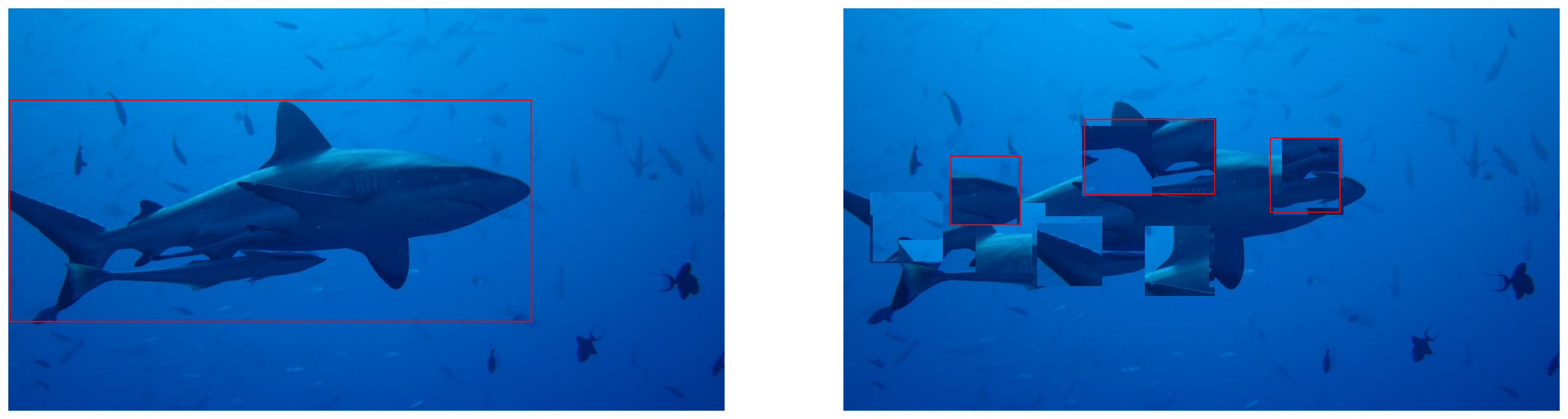}
\end{center}
   \caption{\textbf{Left:} Original image with the detected "airplane" bounding box. \textbf{Right:} Image after random patches within the detected bounding box are swapped. The shark is still misclassified as airplane, which indicates that the model does not make full use of the shape of the object but relies on texture and small subparts of the objects}
\label{fig:long}
\label{fig:onecol}
\label{fig:patch_shuffle_sample}
\end{figure}

\begin{figure}[t]
\begin{center}
   \includegraphics[width=1.0\linewidth]{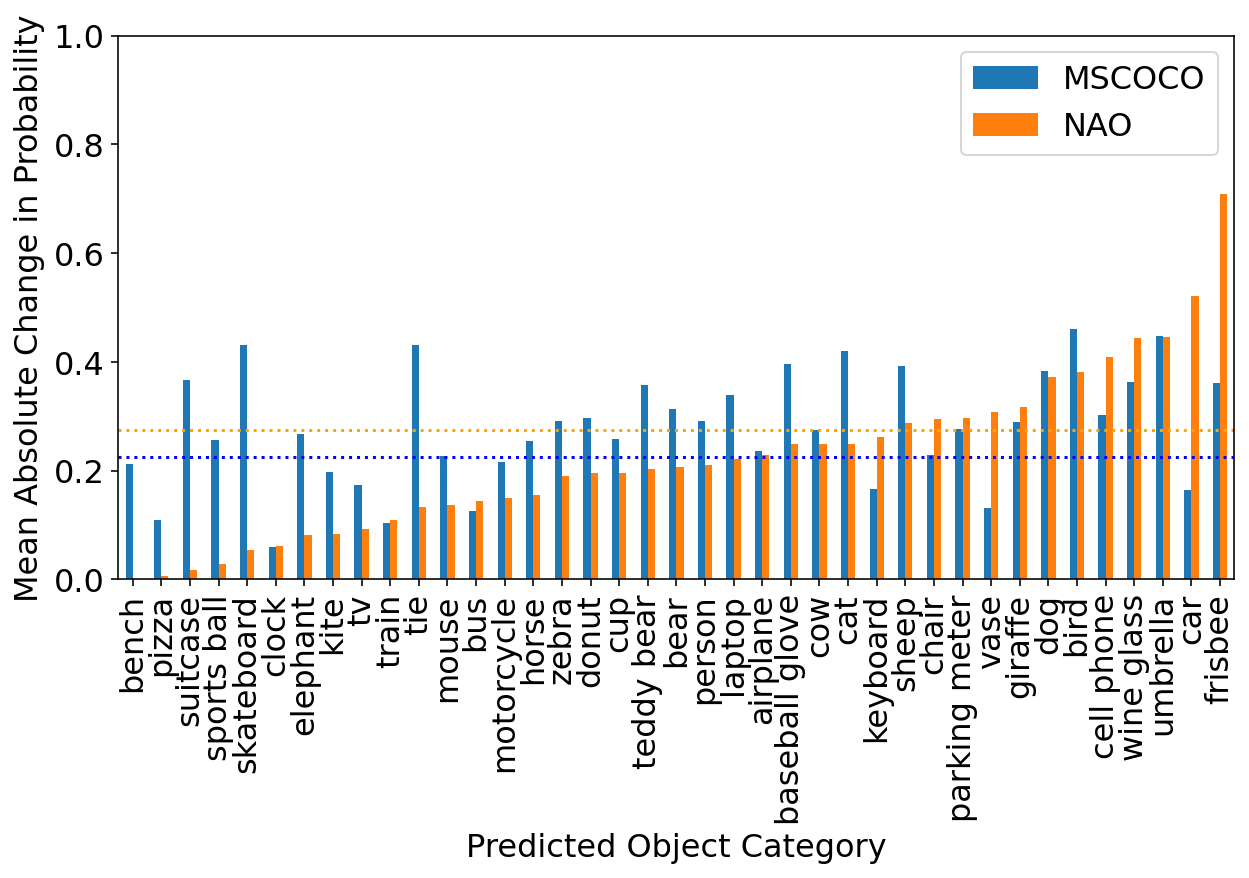}
\end{center}
   \caption{Mean absolute change in probability when patches inside the bounding box are swapped randomly. The blue and orange dotted line represent the mean average change in probability across all object categories for MSCOCO and NAO respectively. This confirms the texture bias hypothesis because even if the shape of the objects are heavily distorted while the local texture is intact, the network is still able to make the same prediction with similar confidence in most object categories.}
\label{fig:long}
\label{fig:onecol}
\label{fig:patch_shuffle_results}
\end{figure}

Geirhos \etal~\cite{geirhos2018imagenettrained} demonstrated that classification networks are biased towards recognizing texture instead of shape. Brendel \etal~\cite{brendel2018approximating} showed that a classification network can still reach a high level of accuracy using just small patches extracted from images. In this work, we show that detection models also show strong local texture bias, making them susceptible to adversarial objects with similar object subparts but are of another object category. For each prediction from EfficientDet-D7 on MSCOCO and NAO, we randomly sample a patch from within the bounding box and swap it with another patch from inside the bounding box. We swap these random patches 3 times such that object is barely recognizable by just the shape. We then match the detected bounding box from the shuffled images to the original bounding box with the highest overlap. We repeat this shuffling process independently 10 times and record the absolute change in probability of the detected object. Figure \ref{fig:patch_shuffle_results} shows that there is only a modest reduction in probability after the shuffles. In Figure \ref{fig:patch_shuffle_sample}, we show an example image where the detection model still misclassifies a shark as an airplane despite patch shuffling.

\section{Conclusion}

We introduce "Natural Adversarial Objects" (NAO), a challenging robustness evaluation dataset for detection models trained on MSCOCO. We evaluated seven state-of-the-art detection models from various families, and show that they consistently fail to perform accurately on NAO, comparing to MSCOCO \textit{val} and \textit{test-dev} set, including on both in-distribution and out-of-distribution objects. We explained the procedure of creating such a dataset which can be useful for creating similar datasets in the future.

We expose that these naturally adversarial objects are difficult to classify correctly due to the "blind-spots" in the MSCOCO dataset. We also utilize integrated gradients, background replacement, and patch shuffling to demonstrate that detection models are overly sensitive to local texture but insensitive to background change, leading such models to be susceptible to natural adversarial objects.


{\small
\bibliographystyle{ieee_fullname}
\bibliography{egbib}

\begin{thebibliography}{10}\itemsep=-1pt

\bibitem{pmlr-v70-arpit17a}
Devansh Arpit, Stanis{\l}aw Jastrz{\k{e}}bski, Nicolas Ballas, David Krueger,
  Emmanuel Bengio, Maxinder~S. Kanwal, Tegan Maharaj, Asja Fischer, Aaron
  Courville, Yoshua Bengio, and Simon Lacoste-Julien.
\newblock A closer look at memorization in deep networks.
\newblock volume~70 of {\em Proceedings of Machine Learning Research}, pages
  233--242, International Convention Centre, Sydney, Australia, 06--11 Aug
  2017. PMLR.

\bibitem{brendel2018approximating}
Wieland Brendel and Matthias Bethge.
\newblock Approximating {CNN}s with bag-of-local-features models works
  surprisingly well on imagenet.
\newblock In {\em International Conference on Learning Representations}, 2019.

\bibitem{7958570}
N. {Carlini} and D. {Wagner}.
\newblock Towards evaluating the robustness of neural networks.
\newblock In {\em 2017 IEEE Symposium on Security and Privacy (SP)}, pages
  39--57, 2017.

\bibitem{s.2018stochastic}
Guneet~S. Dhillon, Kamyar Azizzadenesheli, Jeremy~D. Bernstein, Jean Kossaifi,
  Aran Khanna, Zachary~C. Lipton, and Animashree Anandkumar.
\newblock Stochastic activation pruning for robust adversarial defense.
\newblock In {\em International Conference on Learning Representations}, 2018.

\bibitem{geirhos2018imagenettrained}
Robert Geirhos, Patricia Rubisch, Claudio Michaelis, Matthias Bethge, Felix~A.
  Wichmann, and Wieland Brendel.
\newblock Imagenet-trained {CNN}s are biased towards texture; increasing shape
  bias improves accuracy and robustness.
\newblock In {\em International Conference on Learning Representations}, 2019.

\bibitem{gilmer2018motivating}
Justin Gilmer, Ryan~P Adams, Ian Goodfellow, David Andersen, and George~E Dahl.
\newblock Motivating the rules of the game for adversarial example research.
\newblock {\em arXiv preprint arXiv:1807.06732}, 2018.

\bibitem{Goodfellow2014-rh}
Ian~J Goodfellow, Jonathon Shlens, and Christian Szegedy.
\newblock Explaining and harnessing adversarial examples.
\newblock Dec. 2014.

\bibitem{he2017mask}
Kaiming He, Georgia Gkioxari, Piotr Doll{\'a}r, and Ross Girshick.
\newblock Mask r-cnn.
\newblock In {\em Proceedings of the IEEE international conference on computer
  vision}, pages 2961--2969, 2017.

\bibitem{Hendrycks2019-cn}
Dan Hendrycks, Kevin Zhao, Steven Basart, Jacob Steinhardt, and Dawn Song.
\newblock Natural adversarial examples.
\newblock July 2019.

\bibitem{hochreiter98}
Sepp Hochreiter.
\newblock The vanishing gradient problem during learning recurrent neural nets
  and problem solutions.
\newblock {\em International Journal of Uncertainty, Fuzziness and
  Knowledge-Based Systems}, 6:107--116, 04 1998.

\bibitem{ilyas2019adversarial}
Andrew Ilyas, Shibani Santurkar, Dimitris Tsipras, Logan Engstrom, Brandon
  Tran, and Aleksander Madry.
\newblock Adversarial examples are not bugs, they are features.
\newblock In {\em Advances in Neural Information Processing Systems}, pages
  125--136, 2019.

\bibitem{DBLP:journals/corr/abs-1711-11561}
Jason Jo and Yoshua Bengio.
\newblock Measuring the tendency of cnns to learn surface statistical
  regularities.
\newblock {\em CoRR}, abs/1711.11561, 2017.

\bibitem{Kolesnikov2019-qu}
Alexander Kolesnikov, Lucas Beyer, Xiaohua Zhai, Joan Puigcerver, Jessica Yung,
  Sylvain Gelly, and Neil Houlsby.
\newblock Big transfer ({BiT)}: General visual representation learning.
\newblock Dec. 2019.

\bibitem{46638}
Alex Kurakin, Dan Boneh, Florian Tramèr, Ian Goodfellow, Nicolas Papernot, and
  Patrick McDaniel.
\newblock Ensemble adversarial training: Attacks and defenses.
\newblock 2018.

\bibitem{Kuznetsova2018-bk}
Alina Kuznetsova, Hassan Rom, Neil Alldrin, Jasper Uijlings, Ivan Krasin, Jordi
  Pont-Tuset, Shahab Kamali, Stefan Popov, Matteo Malloci, Alexander
  Kolesnikov, Tom Duerig, and Vittorio Ferrari.
\newblock The open images dataset v4: Unified image classification, object
  detection, and visual relationship detection at scale.
\newblock Nov. 2018.

\bibitem{lin2014microsoft}
Tsung-Yi Lin, Michael Maire, Serge Belongie, James Hays, Pietro Perona, Deva
  Ramanan, Piotr Doll{\'a}r, and C~Lawrence Zitnick.
\newblock Microsoft coco: Common objects in context.
\newblock In {\em European conference on computer vision}, pages 740--755.
  Springer, 2014.

\bibitem{McInnes2018-ph}
Leland McInnes, John Healy, and James Melville.
\newblock {UMAP}: Uniform manifold approximation and projection for dimension
  reduction.
\newblock Feb. 2018.

\bibitem{DBLP:journals/corr/PapernotMG16}
Nicolas Papernot, Patrick~D. McDaniel, and Ian~J. Goodfellow.
\newblock Transferability in machine learning: from phenomena to black-box
  attacks using adversarial samples.
\newblock {\em CoRR}, abs/1605.07277, 2016.

\bibitem{DBLP:journals/corr/PapernotMWJS15}
Nicolas Papernot, Patrick~D. McDaniel, Xi Wu, Somesh Jha, and Ananthram Swami.
\newblock Distillation as a defense to adversarial perturbations against deep
  neural networks.
\newblock {\em CoRR}, abs/1511.04508, 2015.

\bibitem{Recht2019-pm}
Benjamin Recht, Rebecca Roelofs, Ludwig Schmidt, and Vaishaal Shankar.
\newblock Do {ImageNet} classifiers generalize to {ImageNet}?
\newblock Feb. 2019.

\bibitem{redmon2017yolo9000}
Joseph Redmon and Ali Farhadi.
\newblock Yolo9000: better, faster, stronger.
\newblock In {\em Proceedings of the IEEE conference on computer vision and
  pattern recognition}, pages 7263--7271, 2017.

\bibitem{Redmon2018-db}
Joseph Redmon and Ali Farhadi.
\newblock {YOLOv3}: An incremental improvement.
\newblock Apr. 2018.

\bibitem{ren2015faster}
Shaoqing Ren, Kaiming He, Ross Girshick, and Jian Sun.
\newblock Faster r-cnn: Towards real-time object detection with region proposal
  networks.
\newblock In {\em Advances in neural information processing systems}, pages
  91--99, 2015.

\bibitem{russakovsky2015imagenet}
Olga Russakovsky, Jia Deng, Hao Su, Jonathan Krause, Sanjeev Satheesh, Sean Ma,
  Zhiheng Huang, Andrej Karpathy, Aditya Khosla, Michael Bernstein, et~al.
\newblock Imagenet large scale visual recognition challenge.
\newblock {\em International journal of computer vision}, 115(3):211--252,
  2015.

\bibitem{samangouei2018defensegan}
Pouya Samangouei, Maya Kabkab, and Rama Chellappa.
\newblock Defense-{GAN}: Protecting classifiers against adversarial attacks
  using generative models.
\newblock In {\em International Conference on Learning Representations}, 2018.

\bibitem{Sermanet2013-ob}
Pierre Sermanet, David Eigen, Xiang Zhang, Michael Mathieu, Rob Fergus, and
  Yann LeCun.
\newblock {OverFeat}: Integrated recognition, localization and detection using
  convolutional networks.
\newblock pages 1--15, Dec. 2013.

\bibitem{song2018pixeldefend}
Yang Song, Taesup Kim, Sebastian Nowozin, Stefano Ermon, and Nate Kushman.
\newblock Pixeldefend: Leveraging generative models to understand and defend
  against adversarial examples.
\newblock In {\em International Conference on Learning Representations}, 2018.

\bibitem{Sundararajan2017-jl}
Mukund Sundararajan, Ankur Taly, and Qiqi Yan.
\newblock Axiomatic attribution for deep networks.
\newblock Mar. 2017.

\bibitem{Szegedy2013-jl}
Christian Szegedy, Wojciech Zaremba, Ilya Sutskever, Joan Bruna, Dumitru Erhan,
  Ian Goodfellow, and Rob Fergus.
\newblock Intriguing properties of neural networks.
\newblock Dec. 2013.

\bibitem{Tan2019-sq}
Mingxing Tan and Quoc~V Le.
\newblock {EfficientNet}: Rethinking model scaling for convolutional neural
  networks.
\newblock May 2019.

\bibitem{tan2020efficientdet}
Mingxing Tan, Ruoming Pang, and Quoc~V Le.
\newblock Efficientdet: Scalable and efficient object detection.
\newblock In {\em Proceedings of the IEEE/CVF Conference on Computer Vision and
  Pattern Recognition}, pages 10781--10790, 2020.

\end{thebibliography}
}

\end{document}